# Censored Exploration and the Dark Pool Problem

Kuzman Ganchev, Michael Kearns, Yuriy Nevmyvaka, Jennifer Wortman Vaughan
Computer and Information Science, University of Pennsylvania

## Abstract

We introduce and analyze a natural algorithm for *multi-venue exploration from censored data*, which is motivated by the *Dark Pool Problem* of modern quantitative finance. We prove that our algorithm converges in polynomial time to a near-optimal allocation policy; prior results for similar problems in stochastic inventory control guaranteed only asymptotic convergence and examined variants in which each venue could be treated independently. Our analysis bears a strong resemblance to that of efficient exploration/exploitation schemes in the reinforcement learning literature. We describe an extensive experimental evaluation of our algorithm on the Dark Pool Problem using real trading data.

## 1 Introduction

We analyze a framework and algorithm for the problem of *multi-venue exploration from censored data*. Consider a setting in which at each time period, we have some *volume* of $V$ units (possibly varying with time) of an abstract good. Our goal is to "sell" or "consume" as many of these units as possible at each step, and there are $K$ abstract "venues" in which this selling or consumption may occur. We can divide our $V$ units in any way we like across the venues in service of this goal. Our interest in this paper is in how to *efficiently learn* a near-optimal allocation policy over time, under stochastic assumptions on the venues.

This setting belongs to a broad class of problems known in the operations research literature as *perishable inventory problems* (see Related Work below). In the *Dark Pool Problem* (discussed extensively in Section 5), at each time step a trader must buy or sell up to $V$ shares of a given stock on behalf of a client [1], and does so by distributing or allocating them over multiple distinct exchanges (venues) known as *dark pools*. Dark pools are a recent type of stock exchange in which relatively little information is provided about the current outstanding orders (Wikipedia, 2009, Bogoslaw, 2007). The trader would like to execute as many of the $V$ shares as possible. If $v_i$ shares are allocated to dark pool $i$, and all of them are executed, the trader learns only that the liquidity available at exchange $i$ was *at least* $v_i$, not the actual larger number that *could* have executed there; this important aspect of our framework is known as *censoring* in the statistics literature.

In this work we make the natural and common assumption that the maximum amount of consumption available in venue $i$ at each time step (e.g., the total liquidity available in the example above) is drawn according to a fixed but unknown distribution $P_i$. Formally speaking, this means that when $v_i$ units are submitted to venue $i$, a value $s_i$ is drawn randomly from $P_i$ and the observed (and possibly censored) amount of consumption is $\min\{s_i, v_i\}$.

A learning algorithm receives a sequence of volumes $V^1, V^2, \ldots$ and must decide how to distribute the $V^t$ units across the venues at each time step $t$. Our goal is to *efficiently* (in time polynomial in the "complexity" of the $P_i$ and other parameters) learn a near-optimal allocation policy. There is a distinct *between-venue exploration* component to this problem, since the "right" number of shares to submit to venue $i$ may depend on both $V^t$ and the distributions for the *other* venues, and the only mechanism by which we can discover the distributions is by submitting allocations. If we routinely submit too-small volumes to a venue, we receive censored observations and are underutilizing the venue; if we submit too-large volumes we receive uncensored (or *direct*) observations but have excess inventory.

---

[1]In our setting it is important that we view $V$ as given exogenously by the client and not under the trader's control, which distinguishes our setting somewhat from prior works; see Related Work.

Our main theoretical contribution is a provably polynomial-time algorithm for learning a near-optimal policy for any unknown venue distributions $P_i$. This algorithm takes a particularly natural and appealing form, in which *allocation* and distribution *reestimation* are repeatedly alternated. More precisely, at each time step we maintain distributional estimates $\hat{P}_i$; pretending that these estimates are in fact exactly correct, we allocate the current volume $V$ accordingly. These allocations generate observed consumptions in each venue, which in turn are used to update or reestimate the $\hat{P}_i$.

We show that when the $\hat{P}_i$ are "optimistic tail modifications" of the classical *Kaplan-Meier* maximum likelihood estimator for censored data, this estimate-allocate loop has *provably efficient between-venue exploration* behavior that yields the desired result. Venues with smaller available volumes (relative to the overall volume $V^t$ and the other venues) are gradually given smaller allocations in the estimate-allocate loop, whereas venues with repeated censored observations are gradually given larger allocations, eventually settling on a near-optimal overall allocation distribution. Interestingly, the analysis of our algorithm bears strong resemblance to the exploration-exploitation arguments common in the E$^3$ and RMAX family of algorithms for reinforcement learning (Kearns and Singh, 2002, Brafman and Tennenholtz, 2003).

### 1.1 Related Work

The problem perhaps closest to our setting is the widely studied *newsvendor problem* from the operations research literature. In this problem, at each time period a player (representing a newsstand owner) *chooses* the quantity $V$ of newspapers to purchase at a fixed per-unit price, and tries to optimize profit in the face of demand uncertainty at a *single* venue (their newsstand). There is a large and diverse literature on this single-venue problem; see Huh et al. (2009) and the citations within. In this same paper, the authors are the first to consider the use of the Kaplan-Meier estimator in perishable inventory problems. They use an estimate-allocate loop similar to ours, and show *asymptotic* convergence to near-optimal behavior in a single venue. Managing the distribution of an *exogenously specified* volume $V$ across *multiple* venues (which are the important aspects of the Dark Pool Problem, where the volume to be traded is specified by a client, and there are many dark pools), and the attendant exploration-exploitation trade-off *between venues*, are key aspects and differentiators of our algorithm and analysis. We also obtain stronger (polynomial time rather than asymptotic) bounds, which requires a modification of the classical Kaplan-Meier estimator.

Our main theoretical contribution is thus the development and analysis of a multiple venue, polynomial time, near-optimal allocation learning algorithm, while our main experimental contribution is the application of this algorithm to the Dark Pool Problem.

## 2 Preliminaries

We consider the following problem. At each time $t$, a learner is presented with a quantity or *volume* $V^t \in \{1, \cdots, V\}$ of units, where $V^t$ is sampled from an unknown distribution $Q$. The learner must decide on an *allocation* $\vec{v}^t$ of these shares to a set of $K$ known *venues*, with $v_i^t \in \{0, \cdots, V^t\}$ for each $i \in \{1, \cdots, K\}$, and $\sum_{i=1}^{K} v_i^t = V^t$. The learner is then told the number of units $r_i^t$ *consumed* at each venue $i$. Here $r_i^t = \min\{s_i^t, v_i^t\}$, where $s_i^t$ is the maximum consumption level of venue $i$ at time $t$, which is sampled independently from a fixed but unknown distribution $P_i$. If $r_i^t = v_i^t$, we say that the algorithm receives a *censored observation* because it is possible to infer only that $r_i^t \leq s_i^t$. If $r_i^t < v_i^t$, we say that the algorithm receives a *direct observation* because it must be the case that $r_i^t = s_i^t$.

The goal of the learner is to discover a near-optimal one-step allocation policy, that is, an allocation policy that approximately optimizes the expected number of units out of $V^t$ consumed at each time step $t$. (We briefly discuss other objectives at the end of Section 4.4.)

Throughout the remainder of the paper, we use the shorthand $T_i$ for the *tail probabilities* associated with $P_i$. That is, $T_i(s) = \sum_{s' \geq s} P_i(s')$. Clearly $T_i(0) = 1$ for all $i$. We use $\hat{T}_i^t(s)$ for an empirical estimate of $T_i(s)$ at time $t$, and define $\hat{P}_i^t(s) = \hat{T}_i^t(s) - \hat{T}_i^t(s+1)$ to be the empirical estimate of $P_i(s)$ at time $t$.

## 3 Greedy Allocation is Optimal

In this section, we show that given estimates $\hat{T}_i$ of the tail probabilities $T_i$ for each venue $i$, a simple greedy allocation algorithm can to maximize the (estimated) expected number of units consumed at a single time step. The greedy algorithm allocates one unit at a time. The venue to which the next unit is allocated is chosen to maximize the estimated probability that the unit will be consumed; if $v_i$ units have already been allocated to venue $i$, then the estimated probability that the next allocated unit will be consumed is simply $\hat{T}_i(v_i + 1)$. A formal description is given as Algorithm 1 below.

**Theorem 1** *The allocation returned by* Greedy *maximizes the expected number of units consumed in a sin-*

**Algorithm 1**: Optimal allocation algorithm *Greedy*.
**Input**: Volume $V$, tail probability estimates $\{\hat{T}_i\}_{i=1}^K$
**Output**: An allocation $\vec{v}$
$\vec{v} \leftarrow \vec{0}$;
**for** $\ell \leftarrow 1$ **to** $V$ **do**
    $i \leftarrow \mathrm{argmax}_i\, \hat{T}_i(v_i + 1)$;
    $v_i \leftarrow v_i + 1$ ;
**end**
**return** $\vec{v}$

*gle time step, where the expectation is taken with respect to the estimated tail probabilities* $\{\hat{T}_i\}_{i=1}^K$.

**Proof:** Using the fact that tail probabilities must satisfy $\hat{T}_i(s) \geq \hat{T}_i(s')$ for all $s \leq s'$, it it is easy to verify that by greedily adding units to the venues in decreasing order of $\hat{T}_i(s)$, Algorithm 1 returns

$$\mathrm{argmax}_{\vec{v}} \sum_{i=1}^K \sum_{s=1}^{v_i} \hat{T}_i(s)\ \ s.t. \sum_{i=1}^N v_i = V.$$

It remains to show that the expression above is equivalent to the expected number of units consumed. We do this algebraically. For an arbitrary venue $i$

$$\begin{aligned}
&\mathrm{E}_{s \sim \hat{P}_i}[\min(s, v_i)] \\
&= \sum_{s=1}^{\infty} \hat{P}_i(s) \min(s, v_i) = \sum_{s=1}^{v_i-1} s\hat{P}_i(s) + v_i \hat{T}_i(v_i) \\
&= \sum_{s=1}^{v_i-2} s\hat{P}_i(s) + (v_i - 1)\hat{T}_i(v_i - 1) + \hat{T}_i(v_i) \\
&= \hat{T}_i(1) + \ldots + \hat{T}_i(v_i - 1) + \hat{T}_i(v_i) = \sum_{s=1}^{v_i} \hat{T}_i(s).
\end{aligned}$$

The last two lines follow from the observation that for any $s$, $\hat{P}_i(s-1) + \hat{T}_i(s) = \hat{T}_i(s-1)$ and so $(s-1)\hat{P}_i(s-1) + s\hat{T}_i(s) = (s-1)\hat{T}_i(s-1) + \hat{T}_i(s)$. Thus $\sum_{i=1}^K \sum_{s=1}^{v_i} \hat{T}_i(s) = \sum_{i=1}^K \mathrm{E}_{s \sim \hat{P}_i}[\min(s, v_i)]$, which is the expected number of units consumed. ∎

## 4 Censored Exploration Algorithm

In this section we present our main theoretical result, which is a polynomial-time, near-optimal algorithm for multi-venue exploration from censored data. We first provide an overview of the algorithm and its analysis before diving into the technical details. As mentioned in the Introduction, the analysis bears strong resemblance to the "known and unknown state" exploration-exploitation arguments common in the E$^3$ and RMAX algorithms for reinforcement learning (Kearns and Singh, 2002, Brafman and Tennenholtz, 2003).

At the highest level, the algorithm is quite simple and natural. The algorithm maintains estimates $\hat{T}_i^t$ for the true unknown tail probabilities $T_i$ for each venue $i$. These estimates "improve" with time in a particular quantifiable sense which drives the between-venue exploration discussed in the Introduction. At any given time $t$, the current volume $V^t$ is allocated across the venues by simply calling the optimal greedy allocation scheme (Algorithm 1) on the current set of estimated tail probabilities $\hat{T}_i^t$. This results in new censored observations from each venue, which in turn are used to update the estimates $\hat{T}_i^{t+1}$ used at the next time step. Thus the algorithm, which is given as Algorithm 2, implements a continuous allocate-reestimate loop.

**Algorithm 2**: Main algorithm.
**Input**: Volume sequence $V^1, V^2, V^3, \ldots$
Arbitrarily initialize $\hat{T}_i^1$ for each $i$;
**for** $t \leftarrow 1, 2, 3, \ldots$ **do**
    % Allocation Step:
    $\vec{v}^t \leftarrow Greedy(V^t, \hat{T}_1^t, \ldots, \hat{T}_K^t)$;
    **for** $i \in \{1, \ldots, K\}$ **do**
        Submit $v_i^t$ units to venue $i$;
        Let $r_i^t$ be the number of shares sold;
        % Reestimation Step:
        $\hat{T}_i^{t+1} \leftarrow OptimisticKM(\{(v_i^\tau, r_i^\tau)\}_{\tau=1}^t)$;
    **end**
**end**

The only undetermined part of Algorithm 2 is the subroutine *OptimisticKM*, which specifies how we estimate $\hat{T}_i^t$ from the observed data. The most natural choice would be the maximum likelihood estimator on the data. This estimator is well-known in the statistics literature as the *Kaplan-Meier* estimator. In the following section, we describe Kaplan-Meier and derive a new convergence result that suits our particular needs. This result in turn lets us define an "optimistic tail modification" of Kaplan-Meier that becomes our choice for *OptimisticKM*.

The analysis of Algorithm 2, which is developed in detail over the next few sections, proceeds as follows:

**Step 1:** We first review the Kaplan-Meier maximum likelihood estimator for censored data and provide a new finite sample convergence bound for this estimator. This bound allows us to define a "cut-off point" for each venue $i$ such that the Kaplan-Meier estimate of the tail probability $T_i(s)$ for every value of $s$ up to the cut-off point is guaranteed to be "close to" the true tail probability. We then define a slightly modified version of the Kaplan-Meier estimates in which the tail probability of the next unit above the cut-off is modified in an optimistic manner. We show that in conjunction with the greedy allocation algorithm, this minor modification leads to increased exploration, since the next unit beyond the cut-off point always looks at least as good as the cut-off point itself.

**Step 2:** We next prove our main *Exploitation Lemma* (Lemma 4). This lemma shows that at any time step, if it is the case that the number of units allocated to each venue by the greedy algorithm is strictly below the cut-off point for that venue — which can be thought of as being in a *known state* in the parlance of $E^3$ and RMAX — then the allocation is provably $\epsilon$-optimal.

**Step 3:** We then prove our main *Exploration Lemma* (Lemma 5), which shows that on any time step at which the allocation made by the greedy algorithm is *not* $\epsilon$-optimal, it is possible to lower bound the probability that the algorithm explores. Thus, as with $E^3$ and RMAX, anytime we are not in a known state and thus cannot ensure optimal allocation, we are instead assured of exploring.

**Step 4:** Finally, we show that on any sufficiently long sequence of time steps (where "sufficiently long" is polynomial in $K$, $V$, $1/\epsilon$, and $\ln(1/\delta)$, where $\delta$ is a standard confidence parameter), it must be the case that either the algorithm achieves an $\epsilon$-optimal solution on at least a $1 - \epsilon$ fraction of the sequence, or the algorithm has explored sufficiently often to learn accurate estimates of the tail distributions out to $V$ units on every venue. In either case, we can show that with probability $1 - \delta$, at the end of the sequence, the current algorithm achieves an $\epsilon$-optimal solution with probability at least $1 - \epsilon$.

### 4.1 Convergence of Kaplan-Meier Estimators

We begin by describing the standard Kaplan-Meier maximum likelihood estimator for censored data (Kaplan and Meier, 1958, Peterson, 1983), restricting our attention to a single venue $i$. Let $z_{i,s}$ be the true probability that the demand in this venue is *exactly* $s$ units given that the demand is *at least* $s$ units. Using the fact that $1 - z_{i,s}$ is the conditional probability of there being a demand of at least $s$ given that the demand is at least $s - 1$, it is easy to verify that for any $s > 0$, $T_i(s) = \prod_{s'=0}^{s-1}(1 - z_{i,s'})$. At a high level, we can think of Kaplan-Meier as first computing an estimate of $z_{i,s}$ for each $s$ and then using these estimates to compute an estimate of $T_i(s)$.

Let I be an indicator function taking on the value 1 if its input is true and 0 otherwise. Let $D_{i,s}^t = \sum_{\tau=1}^{t} I[r_i^\tau = s, v_i^\tau > s]$ be the number of direct observations of $s$ units up to time $t$, and let $N_{i,s}^t = \sum_{\tau=1}^{t} I[r_i^\tau \geq s, v_i^\tau > s]$ be the number of (direct or censored) observations of at least $s$ units on time steps at which more than $s$ units were requested. The quantity $N_{i,s}^t$ is then the number of times there was an opportunity for a direct observation of $s$ units, whether or not one occurred.

We can then naturally define $\hat{z}_{i,s}^t = D_{i,s}^t / N_{i,s}^t$, with $\hat{z}_{i,s}^t = 0$ if $N_{i,s}^t = 0$. This quantity is simply the empirical probability of a direct observation of $s$ units given that a direct observation of $s$ units was possible. The Kaplan-Meier estimator of the tail probability for any $s > 0$ after $t$ time steps can then be expressed as

$$\hat{T}_i^t(s) = \prod_{s'=0}^{s-1} \left(1 - \hat{z}_{i,s'}^t\right), \quad (1)$$

with $\hat{T}_i^t(0) = T_i(0) = 1$ for all $t$.

Previous work has established convergence rates for the Kaplan-Meier estimator to the true underlying distribution in the case that the submission sequence $v_i^1, \ldots, v_i^t$ is i.i.d. (see, for example, Foldes and Rejto (1981)), and asymptotic convergence for non-i.i.d. settings (Huh et al., 2009). We are not in the i.i.d. case, since the submitted volumes at one venue are a function of the entire history of allocations and executions across all venues. In the following theorem we give a new finite sample convergence bound applicable to our setting.

**Theorem 2** *Let $\hat{T}_i^t$ be the Kaplan-Meier estimate of $T_i$ as given in Equation 1. For any $\delta > 0$, with probability at least $1 - \delta$, for every $s \in \{1, \cdots, V\}$,*

$$\left|T_i^t(s) - \hat{T}_i^t(s)\right| \leq s\sqrt{2\ln(2V/\delta)/N_{i,s-1}^t}.$$

The proof depends on the next lemma, proof omitted.

**Lemma 1** *For each $i \in \{1, \cdots, \ell\}$, let $x_i$ and $y_i$ be real numbers in $[0,1]$, with $|x_i - y_i| \leq \epsilon_i$ for some $\epsilon_i > 0$. Then $|\prod_{i=1}^{\ell} x_i - \prod_{i=1}^{\ell} y_i| \leq \sum_{i=1}^{\ell} \epsilon_i$.*

**Proof of Theorem 2:** We will show that $\hat{z}_{i,s}^t$ converges to $z_{i,s}$, and that this implies that the Kaplan-Meier tail probability estimator converges to $T_i(s)$.

Consider a fixed value of $s$. Let $t_n$ be the index $\tau$ of the $n$th time step at which $r_i^\tau \geq s$ and $v_i^\tau > s$. By definition, there are $N_{i,s}^t$ such time steps total. For each $n \in \{1, \cdots, N_{i,s}^t\}$, let $X_n = \sum_{\ell=1}^{n} (z_{i,s} - I[r_i^{t_\ell} = s])$. It is easy to see that the sequence $X_1, \cdots, X_{N_{i,s}^t}$ forms a martingale; for each $n$, we have that $|X_n - X_{n+1}| \leq 1$ and $E[X_{n+1}|X_n] = X_n$. By Azuma's inequality (see, for example, Alon and Spencer (2000)), for any $\gamma$,

$$\Pr\left(\left|X_{N_{i,s}^t}\right| \geq \gamma\right) \leq 2e^{-\gamma^2/(2N_{i,s}^t)}.$$

Noting that $X_{N_{i,s}^t} = N_{i,s}^t(z_{i,s} - \hat{z}_s^t)$ and setting $\gamma = \epsilon N_{i,s}^t$ gives us that $\Pr\left(|z_{i,s} - \hat{z}_{i,s}^t| \geq \epsilon\right) \leq 2e^{-\epsilon^2 N_{i,s}^t/2}$. Setting this equal to $\delta/V$ and applying a union bound gives us that with probability $1 - \delta$, for every $s \in \{0, \cdots, V-1\}$, $|z_{i,s} - \hat{z}_{i,s}^t| \leq \sqrt{2\ln(2V/\delta)/N_{i,s}^t}$.

Assume this holds for all $s$. Then it follows from Lemma 1 that for any particular $s \in \{1, \cdots, V\}$,

$$|T_i(s) - \hat{T}_i^t(s)| \leq \sum_{s'=0}^{s-1} \sqrt{\frac{2\ln(2V/\delta)}{N_{i,s'}^t}} \leq s\sqrt{\frac{2\ln(2V/\delta)}{N_{i,s-1}^t}}.$$

### 4.2 Modifying Kaplan-Meier

In Algorithm 3 we describe the minor modification of Kaplan-Meier necessary for our analysis. As described above (Step 1), the value $c_i^t$ in this algorithm can intuitively be viewed as a "cut-off point" up to which we are guaranteed to have sufficient data to accurately estimate the tail probabilities using Kaplan-Meier. (This is formalized in Lemma 2 below.) Thus for every quantity $s \leq c_i^t$, we simply let $\hat{T}_i^t(s)$ be precisely the Kaplan-Meier estimate as in Equation 1.

However, to promote exploration, we set the value of $\hat{T}_i^t(c_i^t + 1)$ optimistically to the Kaplan-Meier estimate of the tail probability at $c_i^t$ (*not* at $c_i^t + 1$). This optimistic modification is necessary to ensure that the greedy algorithm explores (i.e., has a chance of making progress towards increasing at least one cut-off value) on every time step for which it is not already producing an $\epsilon$-optimal allocation. In particular, suppose that the current greedy solution allocated no more than $c_i^t$ units to any venue $i$ and exactly $c_j^t$ units to some venue $j$. Using the standard Kaplan-Meier tail probability estimates, it could be the case that this allocation is suboptimal (there is no way to know if it would have been better to include unit $c_j^t + 1$ from venue $j$ in place of a unit from another venue since we do not have an accurate estimate of the tail probability for this unit), and yet no exploration is taking place. By optimistically modifying the tail probability $\hat{T}_i^t(c_i^t + 1)$ for each venue, we ensure that no venue remains unexplored simply because the algorithm unluckily observes a low demand a small number of times.

We now formalize the idea of $c_i^t$ as a "cut-off point" up to which the Kaplan-Meier estimates are accurate. In the results that follow, we think of $\epsilon > 0$ and $\delta > 0$ as fixed parameters of the algorithm.[2]

**Lemma 2** *With probability at least $1 - \delta$, for all $s \leq c_i^t$, $|T_i(s) - \hat{T}_i^t(s)| \leq \epsilon/(8V)$.*

---
[2]In particular, $\epsilon$ corresponds to the value $\epsilon$ specified in Theorem 3, and $\delta$ corresponds roughly to that $\delta$ divided by the polynomial upper bound on time steps.

---

**Algorithm 3**: Subroutine *OptimisticKM* for computing modified Kaplan-Meier estimators. For all $s$, let $D_{i,s}^t$ and $N_{i,s}^t$ be defined as above, and assume that $\epsilon > 0$ and $\delta > 0$ are fixed parameters.

**Input**: Observed data $(\{(v_i^\tau, r_i^\tau)\}_{\tau=1}^t)$ for venue $i$
**Output**: Modified Kaplan-Meier estimators for $i$

% Calculate the cut-off:
$c_i^t \leftarrow \max\{s : s = 0 \text{ or } N_{i,s-1}^t \geq 128(sV/\epsilon)^2 \ln(2V/\delta)\};$

% Compute Kaplan-Meier tail probabilities:
$\hat{T}_i^t(0) = 1;$
**for** $s = 1$ **to** $V$ **do**
　　$\hat{T}_i^t(s) \leftarrow \prod_{s'=0}^{s-1}\left(1 - (D_{i,s'}^t/N_{i,s'}^t)\right);$
**end**

% Make the optimistic modification:
**if** $c_i^t < V$ **then**
　　$\hat{T}_i^t(c_i^t + 1) \leftarrow \hat{T}_i^t(c_i^t);$
**return** $\hat{T}_i^t;$

---

**Proof:** It is always the case that $T_i(0) = \hat{T}_i^t(0) = 1$, so the result holds trivially unless $c_i^t > 0$. Suppose this is the case. Notice that $N_{i,s}^t$ is monotonic in $s$, with $N_{i,s}^t \geq N_{i,s'}^t$ whenever $s \leq s'$. Thus by definition of $c_i^t$, for all $s < c_i^t$, $N_{i,s}^t \geq 128(sV/\epsilon)^2 \ln(2V/\delta)$. The lemma then follows immediately from an application of Theorem 2. ∎

Lemma 3 shows that it is also possible to achieve additive bounds on the error of tail probability estimates for quantities $s$ much *bigger* than $c_i^t$ as long as the tail probability at $c_i^t$ is sufficiently small.

**Lemma 3** *If $\hat{T}_i^t(c_i^t) \leq \epsilon/(4V)$ and the high probability event in Lemma 2 holds, then for all $s$ such that $c_i^t < s \leq V$, $|T_i(s) - \hat{T}_i^t(s)| \leq \epsilon/(2V)$.*

**Proof:** For any $s > c_i^t$, it must be the case that $\hat{T}_i^t(s) \leq \hat{T}_i^t(c_i^t) \leq \epsilon/(4V)$. If the high probability event in Lemma 2 holds, then $T_i(s) \leq T_i(c_i^t) \leq \hat{T}_i^t(c_i^t) + \epsilon/(8V) \leq \epsilon/(2V)$. Since both $T_i(s)$ and $\hat{T}_i^t(s)$ are constrained to lie between 0 and $\epsilon/(2V)$, it must be the case that $|T_i(s) - \hat{T}_i^t(s)| \leq \epsilon/(2V)$. ∎

### 4.3 Exploitation and Exploration Lemmas

With these two lemmas in place, we are ready to state our main *Exploitation Lemma* (Step 2), which formalizes the idea that once a sufficient amount of exploration has occurred, the allocation output by the greedy algorithm will be $\epsilon$-optimal. The proof of this lemma is where the requirement that $\hat{T}_i^t(c_i^t + 1)$ be set optimistically becomes important. In particular, because of the optimistic setting of $\hat{T}_i^t(c_i^t + 1)$, we know that if the greedy policy allocates exactly $c_i^t$ units to a venue $i$, it could not gain too much by reallocating additional units from another venue to venue $i$ instead. In this sense, we create a "buffer" above each cut-off, guaranteeing that it is not necessary to continue exploring as long as one of the two conditions in the lemma statement is met for each venue.

**Lemma 4 (Exploitation Lemma)** *Assume that at time $t$, the high probability event in Lemma 2 holds. If for each venue $i$, either (1) $v_i^t \leq c_i^t$, or (2) $\hat{T}_i^t(c_i^t) \leq \epsilon/(4V)$, the difference between the expected number of units consumed under allocation $\vec{v}^t$ and the expected number of units consumed under the optimal allocation is at most $\epsilon$.*

**Proof:** Let $a_1, \cdots, a_K$ be any optimal allocation of the $V^t$ units. Since both $\vec{a}$ and $\vec{v}^t$ are over $V^t$ units, it must be the case that $\sum_{i:a_i > v_i^t} a_i - v_i^t = \sum_{i:v_i^t > a_i} v_i^t - a_i$. We can thus define an arbitrary one-to-one mapping between the units that were allocated to different venues by the algorithm and the optimal allocation. Consider any such pair in this mapping. Let $i$ be the venue to which the unit was allocated by the algorithm, and let $n$ be the index of the unit in that venue. Similarly let $j$ be the venue to which the unit was allocated by the optimal allocation, and let $m$ be the index of the unit in that venue.

If Condition (1) in the lemma statement holds for venue $i$, then by Lemma 2, since $n \leq v_i^t \leq c_i^t$, $\hat{T}_i^t(n) \leq T_i(n) + \epsilon/(8V) \leq T_i(n) + \epsilon/(2V)$. On the other hand, if Condition (2) holds, then by Lemma 3, it is still the case that $\hat{T}_i^t(n) \leq T_i(n) + \epsilon/(2V)$.

Now, if $v_j^t < c_j^t$ holds with strict inequality, then by Lemma 2, $T_j(v_j^t + 1) \leq \hat{T}_j^t(v_j^t + 1) + \epsilon/(2V)$. If $v_j^t = c_j^t$, then $T_j(v_j^t + 1) \leq T_j(c_j^t) \leq \hat{T}_j^t(c_j^t) + \epsilon/(2V) = \hat{T}_j^t(v_j^t + 1) + \epsilon/(2V)$, where the last equality is due to the optimistic setting of $\hat{T}_j^t(c_j^t + 1)$ in the exploration algorithm. Finally, if $v_j^t > c_j^t$, then it must be the case that Condition (2) in the lemma statement holds for venue $j$ and by Lemma 3 it is still the case that $T_j(v_j^t + 1) \leq \hat{T}_j^t(v_j^t + 1) + \epsilon/(2V)$.

Thus in all three cases, since $m > v_j^t$, we have that $T_j(m) \leq T_j(v_j^t + 1) \leq \hat{T}_j^t(v_j^t + 1) + \epsilon/(2V)$. Since the greedy algorithm chose to send unit $n$ to venue $i$ instead of sending an additional unit to venue $j$, it must be the case that $\hat{T}_i^t(n) \geq \hat{T}_j^t(v_j^t + 1)$ and thus $T_i(n) \geq T_j(m) - \epsilon/V$.

Since there are at most $V$ pairs in the matching, and each contributes at most $\epsilon/V$ to the difference in expected units consumed between the optimal allocation and the algorithm's, the difference is at most $\epsilon$. ∎

Note that this bound is tight in the sense that it is possible to construct examples where the difference in expected units consumed is as large as $\epsilon$.

Finally, Lemma 5 presents the main exploration lemma (Step 3), which states that on any time step at which the allocation is *not* $\epsilon$-optimal, the probability of a "useful" observation is lower bounded by $\epsilon/(8V)$.

**Lemma 5 (Exploration Lemma)** *Assume that at time $t$, the high probability event in Lemma 2 holds. If the allocation is not $\epsilon$-optimal, then for some venue $i$, with probability at least $\epsilon/(8V)$, $N_{i,c_i^t}^{t+1} = N_{i,c_i^t}^t + 1$.*

**Proof:** Suppose the allocation is not $\epsilon$-optimal at time $t$. By Lemma 4, it must be the case that there exists some venue $i$ for which $v_i^t > c_i^t$ and $\hat{T}_i^t(c_i^t) > \epsilon/(4V)$. Let $\ell$ be a venue for which this is true. Since $v_\ell^t > c_\ell^t$, it will be the case that $N_{\ell,c_\ell^t}^{t+1} = N_{\ell,c_\ell^t}^t + 1$ as long as $r_\ell^t \geq c_\ell^t$. Since $\hat{T}_\ell^t(c_\ell^t) > \epsilon/(4V)$, Lemma 2 implies that $T_\ell(c_\ell^t) \geq \epsilon/(8V)$, so $r_\ell^t \geq c_\ell^t$ with probability at least $\epsilon/(8V)$. ∎

### 4.4 Putting It All Together

With the exploitation and exploration lemmas in place, we are finally ready to state our main theorem. The full proof is omitted due to lack of space, but we sketch the main ideas below.

**Theorem 3 (Main Theorem)** *For any $\epsilon > 0$ and $\delta > 0$, with probability $1 - \delta$ (over the randomness of draws from $Q$ and $\{P_i\}$), after running for a time polynomial in $K$, $V$, $1/\epsilon$, and $\ln(1/\delta)$, Algorithm 2 makes an $\epsilon$-optimal allocation on each subsequent time step with probability at least $1 - \epsilon$.*

**Proof Sketch:** Suppose that the algorithm runs for $R$ time steps, where $R$ is a polynomial in $K$, $V$, $1/\epsilon$, and $\ln(1/\delta)$ (to be determined later). If it is the case that the algorithm was already $\epsilon$-optimal on a fraction $(1 - \epsilon)$ of the $R$ time steps, then we can argue that the algorithm will continue to be $\epsilon$-optimal on at least a fraction $(1 - \epsilon)$ of future time steps; this is because the algorithm gets better on average over time as more observations are made.

On the other hand, if the algorithm chose sub-optimal allocations on at least a fraction $\epsilon$ of the $R$ time steps, then by Lemma 5, the algorithm must have incremented $N_{i,c_i^t}^t$ for some venue $i$ and current cut-off $c_i^t$ approximately $\epsilon^2 R/(8V)$ times. By definition of the $c_i^t$, it can never be the case that $N_{i,c_i^t}^t$ was incremented too many times for any *fixed* values of $i$ and $c_i^t$ (where "too many" is a polynomial in $V$, $1/\epsilon$, and $\ln(1/\delta)$); otherwise the cut-off would have increased. Since there are only $K$ venues and $V$ possible cut-off values to consider in each venue, the total number of increments can be no more than $KV$ times this polynomial, another polynomial in $V$, $1/\epsilon$, $\ln(1/\delta)$, and now $K$. If $R$ is sufficiently large (but still polynomial in all of the desired quantities) and approximately $\epsilon^2 R/(8V)$ increments were made, we can argue that *every* venue must have been fully explored, in which case, again, future allocations will be $\epsilon$-optimal. ∎

We conclude our theory contributions with a few brief remarks. Our optimistic tail modifications of the Kaplan-Meier estimators are relatively mild [3]. In many circumstances we actually expect that the estimate-allocate loop with unmodified Kaplan-Meier would work well, and we investigate a parametric version of this learning algorithm in the experiments described below.

It also seems quite likely that variants of our algorithm and analysis could be developed for alternative objective functions, such as minimizing the number of steps of repeated reallocation required to execute all shares (for which it is possible to create examples where myopic greedy allocation at each step is suboptimal), or maximizing the probability of executing all $V$ shares on a single step.

## 5 Application: Dark Pool Problem

We now describe the application that provided the original inspiration to develop our framework and algorithm. As mentioned in the Introduction, *dark pools* are a particular and relatively recent type of exchange for listed stocks. While the precise details are beyond the scope of this paper, the main challenge in executing large-volume trades in traditional ("light") exchanges is that it is difficult to "conceal" such trades, and their revelation generally results in adverse impact on price (e.g. the presence of a large-volume buyer causes the price to rise against that buyer). If the volume is sufficiently large, this difficulty of concealment remains even if one attempts to break the trade up slowly over time. Dark pools arose exactly to address the problems faced by large traders, and tend to emphasize trade and order concealment over price optimization (Wikipedia, 2009, Bogoslaw, 2007).

In a typical dark pool, buyers and sellers submit orders that simply specify the total volume of shares they wish to buy or sell, with the price of the transaction determined exogenously by "the market". [4] Upon submitting an order to (say) buy $v$ shares, a trader is put in a queue of buyers awaiting transaction, and there is a similar queue of sell orders. Matching between buyers and sellers occurs in sequential arrival of orders, similar to a light exchange. Unlike a light exchange, no information is provided to traders about how many parties or shares might be available in the pool at any given moment. Thus in a given time period, a submission of $v$ shares results only in a (possibly censored) report of how many shares up to $v$ were executed.

While presenting their own trading challenges, dark pools have become tremendously popular exchanges, responsible for executing 10-20% of the overall US equity volume. In fact, they have been so successful that there are now approximately 40+ dark pools for the U.S. Equity market alone, leading traders and brokerages to face exactly the censored multi-venue exploration problem we have been studying: How should one optimally distribute a large trade over the many independent dark pools? We now describe the application of our algorithm to this problem, basing it on actual data from four active dark pools.

### 5.1 Summary of Dark Pool Data

Our data set is from the internal dark pool order flow for a major U.S. broker-dealer. Each (possibly censored) observation in this data is exactly of the form discussed throughout the paper — a triple consisting of the dark pool name, the number of shares sent to that pool, and the number of shares subsequently executed within a short time interval. It is important to highlight some limitations of the data. First, note that the data set conflates the policy the brokerage used for allocation across the dark pools with the liquidity available in the pools themselves. For our data set, the policy in force was very similar to the bandit-style approach we discuss below. Second, the "parent" orders determining the overall volumes to be allocated across the pools were determined by the brokerage's trading needs, and are similarly out of our control.

The data set contains submissions and executions for four active dark pools: BIDS Trading, Automated Trading Desk, D.E. Shaw, and NYFIX, each for a dozen of relatively actively-traded stocks [5], thus yielding 48 distinct stock-pool data sets. The average daily trading volume of these stocks across all exchanges (light and dark) ranges from 1 to 60 million shares, with a median volume of 15 million shares. Energy, Financials, Consumer, Industrials, and Utilities industries are represented. Our data set spans 30 trading days; for every stock-pool pair we have on average 1,200 orders (from 600 to 2,000), which corresponds to 1.3 million shares (from 0.5 million to 3 million). Individual order sizes range from 100 to 50,000 shares, with 1,000 shares being the median. 16% of orders are filled at least partially (meaning that fully 84% result in no shares executed), 9% of the total submitted volume was executed, and 11% of all observations were censored.

---

[3] The same results can be proved for more invasive modifications, such as pushing all mass above the cut-offs to the maximum possible volume, which would result in even more aggressive exploration.

[4] Typically the midpoint between the bid and ask in the light exchanges. This is a slight oversimplification but accurate for our purposes.

[5] Tickers represented are AIG, ALO, CMI, CVX, FRE, HAL, JPM, MER, MIR, NOV, XOM, and NRG.

### 5.2 Parametric Models for Dark Pools

While the theory and algorithm we have developed for censored exploration permit a general (nonparametric) form for the venue distributions $P_i$, in any application it is reasonable to ask whether the data permits a simple parametric form for these distributions, with the attendant computational and sample complexity benefits. For our dark pool data the answer to this question is affirmative.

We experimented with a variety of common parametric forms for the distributions. For each such form, the basic methodology was the same. For each of the $4 \times 12 = 48$ venue-stock pairs, the data for that pair was split evenly into a training set and a test set. The training data was used to select the maximum likelihood model from the parametric class. Note that we can no longer directly apply the nonparametric Kaplan-Meier estimator — within each model class, we must directly maximize the likelihood on the censored training data. This is a relatively straightforward and efficient computation for each of the model classes we investigated. The test set was of course used to measure the generalization performance of each maximum likelihood model.

Our investigation revealed that the best models maintained a separate parameter for the probability of 0 shares being available (that is, $P_i(0)$ is explicitly estimated) — a "Zero Bin" or ZB parameter. This is due to the fact that the vast majority of submissions (84%) to dark pools result in no shares being executed. We then examined various parametric forms for the non-zero portions of the venue distributions [6], including uniform (which of course requires no additional parameters), and Poisson, exponential and power law forms (each of which requires a single additional parameter).

The generalization results strongly favor the power law form, in which the probability of $s$ shares being available is proportional to $1/s^\beta$ for real $\beta$ — a so-called heavy-tailed distribution when $\beta > 0$. Nonparametric models trained with Kaplan-Meier are best on the training data but overfit badly due to their complexity relative to the sparse data, while the other parametric forms cannot accommodate the heavy tails of the data. The comparative results are summarized in Table 1. Based on this comparison, for our dark pool study we investigate a variant of our main algorithm, in which the estimate-allocate loop has an estimation step using maximum likelihood estimation within the ZB + Power Law model, and allocations are done greedily on these same models.

---

[6] These forms were applied up to the largest volume submitted in the data sets, then normalized.

| Model | Train Loss | Test Loss | Wins |
|---|---|---|---|
| Nonparametric | 0.454 | 0.872 | 3 |
| ZB + Uniform | 0.499 | 0.508 | 12 |
| ZB + Power Law | 0.467 | 0.484 | 28 |
| ZB + Poisson | 0.576 | 0.661 | 0 |
| ZB + Exponential | 0.883 | 0.953 | 5 |

Table 1: Average per-sample log-loss (negative log likelihood) for the different venue distribution models. The "Wins" column counts the number of stock-venue pairs where a given model beats the other four on the test data.

In terms of the estimated ZB + Power Law parameters themselves, we note that for all 48 stock-pool pairs the Zero Bin parameter accounted for most of the distribution (between a fraction 0.67 and 0.96), which is not surprising considering the aforementioned preponderance of entirely unfilled orders in the data. The vast majority of the 48 exponents $\beta$ fell between $\beta = 0.25$ and $\beta = 1.3$ — so rather long tails indeed — but it is noteworthy that for one of the four dark pools, 7 of the 12 estimated exponents were actually *negative*, yielding a model that predicts *higher* probabilities for larger volumes. This is likely an artifact of our size- and time-limited data set, but is not entirely unrealistic and results in some interesting behavior in the simulations below.

### 5.3 Data-Based Simulation Results

As in any control problem, the dark pool data in our possession is unfortunately insufficient to evaluate and compare different allocation algorithms. This is because of the aforementioned fact that the volumes submitted to each venue were fixed by the specific policy that generated the data, and we cannot explore alternative choices — if our algorithm chooses to submit 1000 shares to some venue, but in the data only 500 shares were submitted, we simply cannot infer the outcome of our desired submission.

We thus instead use the raw data to derive a *simulator* with which we can evaluate different approaches. In light of the modeling results of Section 5.2, the simulator for stock $S$ was constructed as follows. For each dark pool $i$, we used *all* of the data for $i$ and stock $S$ to estimate the maximum likelihood Zero Bin +

Power Law distribution. (Note that there is no need for a training-test split here, as we have already separately validated the choice of distributional model.) This results in a set of four venue distribution models $P_i$ that form the simulator for stock $S$. This simulator accepts allocation vectors $(v_1, v_2, v_3, v_4)$ indicating how many shares some algorithm wishes to submit to each venue, draws a "true liquidity" value $s_i$ from $P_i$ for each $i$, and returns the vector $(r_1, r_2, r_3, r_4)$, where $r_i = \min(v_i, s_i)$ is the possibly censored number of shares filled in venue $i$.

Across all 12 stocks, we compared the performance of four different allocation algorithms:

• The *ideal* allocation, which is given the *true parameters* of the ZB + Power Law distributions used by the simulator, and allocates shares optimally (greedily) with respect to these distributions.

• The *uniform* allocation, which always divides any order equally among all four venues.

• Our *learning* algorithm, which implements the repeated allocate-reestimate loop, using the maximum likelihood ZB + Power Law model for the reestimation step.

• A simple *bandit*-style algorithm, which begins with equal weights assigned to all four venues. Allocation to a venue which results in any nonzero number of shares being executed causes that venue's weight to be multiplied by a factor $\alpha > 1$ [7]. Allocations are always proportional to the current weight vector.

Some remarks on these algorithms are in order. First, note that the first two allocation methods (ideal and uniform) are non-adaptive and are meant to serve as baselines — one of them the best performance we could hope for (ideal), and the other the most naive allocation possible (uniform). Second, note that our algorithm has a distinct advantage in the sense that it is using the correct parametric form, the same being used by the simulator itself. Thus our evaluation of this algorithm is certainly optimistic compared to what should be expected "in practice". Finally, note that the bandit algorithm is the crudest type of weight-based allocation scheme of the type that abounds in the no-regret literature (Cesa-Bianchi and Lugosi, 2006); we are effectively forcing our problem into a 0/1 loss setting corresponding to "no shares" and "some shares" being executed. Certainly more sophisticated bandit-style approaches can and should be examined.

Each algorithm was run in simulation for some number of *episodes*. Each episode consisted of the allocation of a fixed number $V$ of shares — thus the same number of shares is repeatedly allocated by the algorithm, though of course this allocation will change over time for the two adaptive algorithms as they learn. Each episode of simulation results in a some fraction of the $V$ shares being executed. Two values of $V$ were investigated — a smaller and therefore easier value $V = 1000$, and the larger and therefore more difficult $V = 8000$.

---

[7] For the experiments we optimized $\alpha$ over all stock-pool pairs and thus used $\alpha = 1.05$.

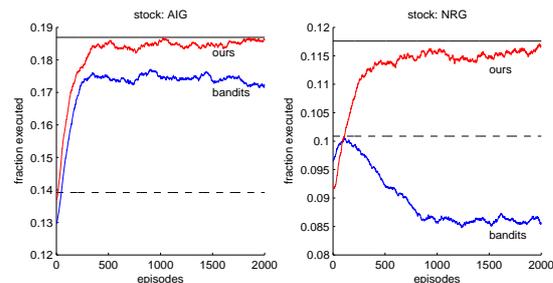

Figure 1: Sample learning curves. For the stock AIG (left panel), the bandits algorithm (labeled blue curve) beats uniform allocation (dashed horizontal line) but appears to asymptote short of ideal allocation (solid horizontal line). For the stock NRG (right panel), the bandits algorithm actually deteriorates with more episodes, underperforming both uniform and ideal allocation. For both these stocks (and for the other 10 in our data set), our algorithm (labeled red curve) performs nearly optimally.

We begin by showing full learning curves over 2000 episodes with $V = 8000$ for a couple of representative stocks in Figure 1. Here the average performance of the two non-adaptive allocation schemes (Ideal and Uniform) are represented as horizontal lines, while learning curves are given for the adaptive schemes. Due to high variance of the heavy-tailed venue distributions used by the simulator, a single trial of 2000 episodes is extremely noisy, so we both average over 400 trials for each algorithm, and smooth the resulting averaged learning curve with a standard exponential decay temporal moving average.

We see that our learning algorithm converges towards the ideal allocation (as suggested by the theory), often relatively quickly. Furthermore, in each case this ideal asymptote is significantly better than the uniform allocation strawman, meaning that optimal allocations are highly non-uniform. Learning curves for the bandit approach exhibit one of three general behaviors over the set of 12 stocks. In some cases, the bandit approach is quite competitive with our algorithm, though converging to ideal perhaps slightly slower (not shown in Figure 1). In other cases, the bandit approach learns to outperform uniform allocation but appears to asymptote short of the ideal allocation. Finally, in some cases the bandit approach appears to actually "learn the wrong thing", with performance decaying significantly with more episodes. This happens when one venue has a very heavy tail, but also a relatively high probability of executing 0 shares, and occurs because the bandit approach does not have an explicit representation of the tails of the distribution.

The left column of Figure 2 shows more systematic head-to-head comparisons of our algorithm's performance versus the ideal, uniform and bandit allocations after 2000 episodes for both small and large $V$. The values plotted are averages of the last 50 points on learning curves similar to Figure 1. These scatterplots show that across all 12 stocks and both settings of $V$, our algorithm competes well with the optimal allocation, dramatically outperforms uniform, and significantly outperforms the bandit allocations (especially at the larger volume $V = 8000$). The average completion rate across all stocks for the large (small) order sequences is 10.0% (13.1%) for uniform and 13.6% (19.4%) for optimal allocations. Our algorithm performs almost as well as optimal — 13.5% (18.7%) — and much better than bandits at 11.9% (17.2%).

The right column of Figure 2 is quite similar, except now we measure performance not by the fraction of $V$ shares filled in one step, but the natural alternative of *order half-life* — the number of steps of *repeated* resubmission of any remaining shares to get the total number executed above $V/2$. Despite the fact that our algorithm is not designed to optimize this criterion and that our theory does not directly apply to it, we see the same broad story on this metric as well — our algorithm competes with ideal, dominates uniform allocation and beats the bandit approach on large orders. The average order half-life for large (small) orders is 7.2 (5.3) for uniform allocation and 5.9 (4.4) for the greedy algorithm on the true distributions. Our algorithm requires on average 6.0 (4.9) steps, while bandits uses 7.0 (4.4) to trade the large (small) orders.

## Acknowledgments

We are grateful to Curtis Pfeiffer and Andrew Westhead for valuable conversations, and to Bobby Kleinberg for introducing us to the literature on the newsvendor problem.

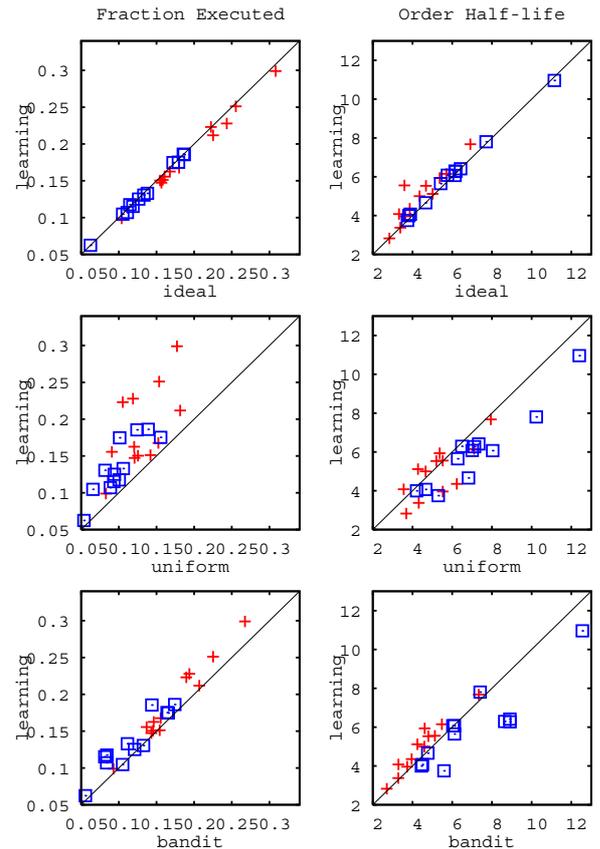

Figure 2: Comparison of our learning algorithm to the three baselines (ideal, uniform, and bandits). In each plot, the performance of the learning algorithm is plotted on the $y$ axis, and the performance of one of the baselines on the $x$ axis. Left column: Evaluated by the fraction of submitted shares executed in a single time step. Here higher values are better, and points above the diagonal are wins for our algorithm. Right: Evaluated by order half-life. Here lower values are better, and points below the diagonal are wins for our algorithm. Each point corresponds to a single stock and order size; small orders (red plus signs) are 1000 shares, large orders (blue squares) are 8000 shares.